\documentclass{article}

\usepackage[preprint]{neurips_2025}

\usepackage[utf8]{inputenc} 
\usepackage[T1]{fontenc}    
\usepackage{hyperref}       
\usepackage{url}            
\usepackage{booktabs}       
\usepackage{amsfonts}       
\usepackage{nicefrac}       
\usepackage{microtype}      
\usepackage{xcolor}         
\usepackage{graphicx}
\usepackage{amsmath}
\usepackage{multirow}
\usepackage{multicol}
\usepackage{colortbl}
\usepackage{tcolorbox}
\usepackage{subfigure}
\usepackage{bm}

\usepackage[numbers]{natbib}

\definecolor{light-gray}{gray}{0.6}
\definecolor{front-color}{HTML}{F5FFFA}
\definecolor{Gray}{gray}{0.93}

\title{Decoupled Seg Tokens Make Stronger Reasoning Video Segmenter and Grounder}
\author{
  Jisheng Dang$^{1,2}$, 
  Xudong Wu$^{3}$,  
  Bimei Wang$^{4,2}$, 
  Ning Lv$^{1}$, 
  Jiayu Chen$^{1}$, \\
  \textbf{Jingwen Zhao$^{3}$},
  \textbf{Yichu liu$^{5}$},
  \textbf{Jizhao Liu$^{1}$},
  \textbf{Juncheng Li$^{6}$},
  \textbf{Teng Wang$^{7}$}\\
  \\
  $^{1}$Lanzhou University, 
  $^{2}$National University of Singapore,
  $^{3}$Sun Yat-sen University,\\
  $^{4}$Jinan University,
  $^{5}$South China University of Technology,\\
  $^{6}$Zhejiang University,
  $^{7}$The University of Hong Kong
  \\
}

\begin{document}

\maketitle

\begin{abstract}
Existing video segmenter and grounder approaches, exemplified by Sa2VA, directly fuse features within segmentation models. This often results in an undesirable entanglement of dynamic visual information and static semantics, thereby degrading segmentation accuracy. To systematically mitigate this issue, we propose DeSa2VA, a decoupling-enhanced prompting scheme integrating text pre-training and a linear decoupling module to address the information processing limitations inherent in SAM-2. Specifically, first, we devise a pre-training paradigm that converts textual ground-truth labels into point-level prompts while generating corresponding text masks. These masks are refined through a hybrid loss function to strengthen the model's semantic grounding capabilities. Next, we employ linear projection to disentangle hidden states that generated by a large language model into distinct textual and visual feature subspaces. Finally, a dynamic mask fusion strategy synergistically combines these decoupled features through triple supervision from predicted text/visual masks and ground-truth annotations. Extensive experiments demonstrate state-of-the-art performance across diverse tasks, including image segmentation, image question answering, video segmentation, and video question answering. Our codes are available at 
\href{https://github.com/longmalongma/DeSa2VA}{https://github.com/longmalongma/DeSa2VA}.
\end{abstract}

\section{Introduction}

Recent advancements in segmentation methodologies, particularly the emergence of foundation models, have significantly transformed the field of computer vision. The Segment Anything Model (SAM)~\cite{kirillov2023segment} introduced promptable segmentation, facilitating zero-shot segmentation of unseen objects via point or pixel prompts through a chain-of-thought mechanism. This paradigm blurs traditional distinctions between segmentation and recognition. Building upon SAM, SAM-2~\cite{ravi2024sam} achieves faster inference, higher accuracy, and improved handling of multidimensional data, extending its applicability to video-level tasks.

 Multimodal Large Language Models (MLLMs) have further advanced the refinement of segmentation tasks by integrating linguistic context with visual processing. Methods such as Sa2VA~\cite{yuan2025sa2va}
 and MemorySAM integrate MLLMs with segmentation models by leveraging language-driven features to guide segmentation or exploit segmentation cues to enhance language understanding. However, a critical challenge remains: SAM-2 inherently relies on point-based prompts, whereas MLLMs generate high-dimensional,  hidden states. Bridging this modality gap to enable effective interpretation of  signals by segmentation models, is crucial for improving accuracy and efficiency.

\begin{figure}[ht]
\centering
\includegraphics[width=\textwidth]{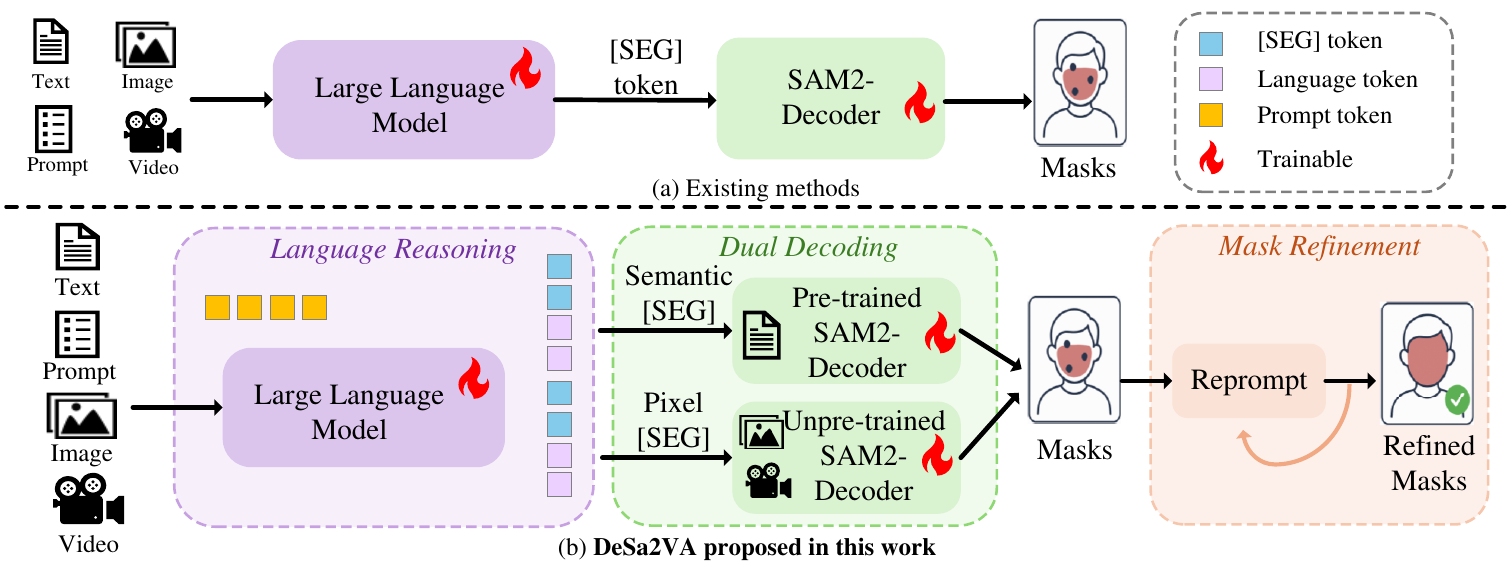}
\caption{ \textbf{Innovation of Our Proposed DeSa2VA.} (a) Baseline model. (b) Our model introduces a decoupling-enhanced prompt module, decoupling  information into textual and visual cues to enhance the segmentation model's prompts. We introduced text understanding pre-training to enable SAM-2 to process decoupled textual information.
}
\label{fig:lambda1}
\end{figure}

Sa2VA tackles this challenge by jointly encoding image-text inputs with an MLLM. 
Segmentation-relevant  representations are tagged with a [SEG] token and fed into SAM-2~\cite{ravi2024sam}, guiding mask generation and enabling scene-level understanding of static/dynamic content. 
Although Sa2VA demonstrates commendable performance in image and video segmentation tasks as well as image and video question answering tasks, three key limitations hinder its effectiveness: \textbf{i) Insufficient Semantic Information.} The [SEG] token lacks rich semantic content, limiting alignment between MLLM outputs and the visual capabilities of SAM-2. \textbf{ii) Limited Textual Understanding in SAM-2.} SAM-2 lacks explicit training for text-based tasks, leading to a misalignment between visual prompts and textual semantics. \textbf{iii) Absence of Fine-Grained Visual Guidance} The [SEG] token provides only high-level hints, restricting the decoder's capacity to generate precise segmentation masks.
To address these limitations, we introduce DeSa2VA (Decoupled Semantic-Aware Visual Augmentation), a novel framework that decouples contextual information while enhancing prompt-based feature learning.
Unlike existing approaches such as InternVL~\cite{chen2024internvl} and SAM-2~\cite{ravi2024sam}, our method explicitly decouples visual and textual modalities to improve prompt quality. By processing modality-specific features separately, the segmentation model achieves a better understanding of linguistic and visual cues, achieving more accurate and robust predictions.  
Specifically, the framework employs a dual-linear-layer architecture to decouple textual and visual information from MLLM outputs. Two parallel linear layers learn modality-specific representations respectively: one captures text-based annotations, while the other extracts visual cues. These layers generate separate hidden states, which are subsequently convolved and passed to SAM-2 to guide mask generation (Fig.~\ref{fig:lambda1}). 

To ensure training stability, text labels are transformed into SAM-2-interpretable point prompts and combined with image pixels to generate supervisory masks. The model learns text-visual alignment by minimizing pixel-wised cross-entropy loss and dice loss between the predicted and ground-truth masks. 
Following the pre-training stage, modality-specific representations are extracted and refined in the decoupling phase. With minimal additional training, these decoupled features support downstream tasks (\emph{e.g.}, visual question answering, reference segmentation).  

In summary, the main contributions of this work are as follows:
\begin{itemize}
    \item We propose a novel  decoupling strategy, which decouples MLLM-generated annotations into distinct, label-rich text/visual representations, enabling efficient use of signals with minimal training overhead.
    \item We introduce a text-visual alignment training, which aligns textual annotations with visual features, training SAM-2 to generate text-grounded masks via supervised loss. 
    \item Our method achieves state-of-the-art results in  segmentation and visual question answering, with ablation studies confirming robust generalization.
\end{itemize}

\section{Related Work}
\label{sec:rel}
\textbf{Multimodal Large Language Models and Referring Segmentation.} Early research on multi-modal fusion focused on feature extractors and fusion strategies for vision-language tasks~\cite{huang2020pixel,li2022blip,li2023blip,dang2025reinforcing,dang2025synpo,dang2025mupa}, with advances in Large Language Models (LLMs)~\cite{brown2020language,touvron2023llama2} leading to multi-modal instruction tuning~\cite{bai2023qwen,chen2023llava,liu2024visual}. This progress has driven the creation of evaluation benchmarks~\cite{fu2023mme,hudson2019gqa} and frameworks such as LLaVA~\cite{liu2024visual}, which integrates tokenization of visual features for visual question answering. Recent developments have extended LLaVA for specialized tasks, including visual grounding and video-based question answering~\cite{dong2024mastering,huang2025reason3d}, while others have unified image and video processing~\cite{li2024llava-onevision}. Parallel advancements in visual perception, like SAM-2~\cite{ravi2024sam}, enable joint image-video interactive segmentation. In referring segmentation, early approaches~\cite{ding2023mevis,khoreva2018video} used specialized fusion modules, while transformer-based methods~\cite{yan2023universal} unified video instance segmentation with object tracking. Recent works~\cite{lai2024lisa,luqi2024generalizable} explore reasoning and dual tasks for referring segmentation and caption generation. Our Sa2VA integrates SAM-2 with contemporary vision-language models, extending these efforts into the video domain to improve performance on both referring segmentation and VQA tasks.

\textbf{Video Segmentation and Grounding.} 
Contemporary video segmentation methods~\cite{hwang2021video,li2023tube-link,zhu2022instance} primarily address closed-set pixel-level segmentation and tracking. While recent work~\cite{guo2023openvis,zhou2023rethinking} explores open-vocabulary settings, their scope remains limited compared to the knowledge capacity of LLMs. In video grounding, studies like~\cite{huang2024vtimelm} leverage LLMs for joint video-audio understanding. VISA~\cite{yan2024visa} investigates reasoning-based video object segmentation but suffers from limited scalability due to task-specific training and lack of end-to-end optimization. To address these gaps, we propose Sa2VA, a model enabling fine-grained spatial-temporal modeling of static (image) and dynamic (video) content, delivering state-of-the-art performance across multiple tasks.


\section{Method}
\label{sec:method}

\subsection{Overview}

The SAM-2 segmentation model \cite{ravi2024sam} supports both sparse (e.g., points, bounding boxes) and dense (e.g., masks) input prompts. Sparse prompts combine point-level inputs with positional encodings and prompt-specific embeddings, while dense prompts are processed with convolutional encoding and fused with frame-level embeddings. Unlike pixel- or mask-level inputs, MLLMs generate higher-level semantic representations that encode cross-modal relationships. Sa2VA reformulates these features into point-based prompts for integration with the segmentation model, but multi-modal data introduces information loss and limits the model's ability to capture complex cross-modal interactions. To address this, we propose a multi-modal decoupling module that separates text and visual modalities, enabling better integration of heterogeneous modalities and improving segmentation accuracy. The architecture decouples the MLLM outputs, using text inputs to train the text encoder and visual data for the original encoder. The combined outputs are processed by the decoder to produce segmentation masks, and after training, the model generates text-based question answering and segmentation results for images and videos.

\begin{figure}[ht]
\
\centering
\includegraphics[width=\textwidth]{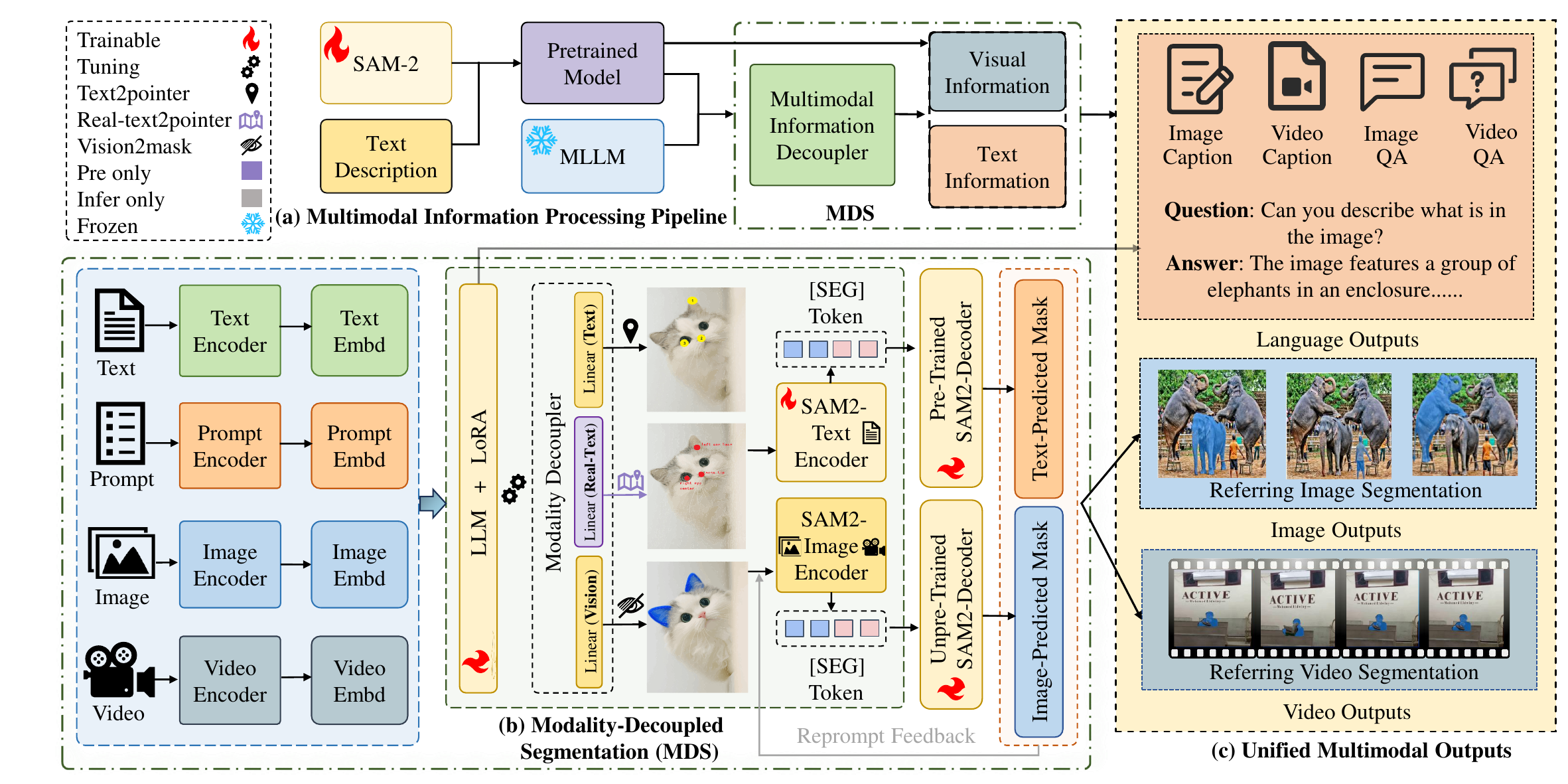}
\caption{\textbf{The Architecture of Our Proposed DeSa2VA Model.} (a) We pre-train SAM-2 to map semantic labels to point-level features via text encoding. (b) Our method separates the outputs of MLLM into visual and textual streams, processed by an untrained SAM-2 decoder and the pre-trained text decoder respectively. (c) The two masks are merged to produce the final output mask, while the MLLM-generated question-answer information is directly output.}
\label{fig:lambda}
\end{figure}

\subsection{Decoupling Text and Visual Modal Information}
\textbf{Linear Layer Projection.}
A linear layer serves as a flexible space that accommodates various dimensions, allowing information from different modalities to be projected into this space. This facilitates the learning of features from data within a unified space, enabling both forward and backward propagation, which enhances the model's ability to comprehend information. In this study, the linear layer is employed to assist in decomposing information.

The information recognized by the segmentation model comes from the large language model InternVL~\cite{chen2024internvl}, which consists of a set of hidden states combining both textual and visual information. In this study, three linear layers are incorporated into our model. Two of these layers are used to decouple the information, projecting the visual and textual features into these two distinct linear layers through self-supervision in the model. The textual feature is denoted as $ \mathbf{x}_{\text{text}} \in \mathbb{R}^{D} $, and the visual feature as $ \mathbf{x}_{\text{vision}} \in \mathbb{R}^{D} $, both originating from the same sample space. The model processes these features through the linear layers to ensure that the textual and visual information share a unified format within the linear layers, enabling the model to handle both types of information in supervised learning. Here, $D$ represents the shared latent space dimension. After projection, the textual representation becomes $ \mathbf{h}_{\text{text}} \in \mathbb{R}^H $, and the visual representation becomes $ \mathbf{h}_{\text{vision}} \in \mathbb{R}^H $, as follows:
\begin{align}
\mathbf{h}_{\text{text}} &= \mathbf{W}_{\text{text}}^\top \mathbf{x}_{\text{text}} + \mathbf{b}_{\text{text}} \quad (\mathbf{W}_{\text{text}} \in \mathbb{R}^{D \times H}) \label{eq:text_proj}, \\
	\mathbf{h}_{\text{vision}} &= \mathbf{W}_{\text{vision}}^\top \mathbf{x}_{\text{vision}} + \mathbf{b}_{\text{vision}} \quad (\mathbf{W}_{\text{vision}} \in \mathbb{R}^{D \times H}) \label{eq:vis_proj}.
\end{align}
The third linear layer is used to process the real textual information, which is employed for training the decoder of the segmentation model. Although the real textual information originates from a different space compared to the visual and textual information discussed above, for subsequent computation, it is necessary to ensure that the information from different modalities and sources retains the same dimensions and shape. Therefore, this study unifies the real textual information with the predicted textual and visual information from the large language model into the same linear layer form. In the first training phase, when training a segmentation model to understand text, the model only uses a linear layer that processes real text information. In the decoupling training phase, all three linear layers are used.

\textbf{Adversarial Training: Modality Discrimination-Driven Disentanglement Game.} To eliminate modality confusion between disentangled features $\bm{h}_t$ and $\bm{h}_v$, an adversarial training framework is proposed and formalized as a min-max optimization problem. This game forces the disentangled features to be indistinguishable from the opposite modality, thereby enhancing the modality separation:
\begin{equation}
    \min_{\theta_t, \theta_v} \max_{\phi_t, \phi_v} \mathbb{E}_{\bm{h}_f} \left[ 
        \log D_{\phi_v}(\bm{h}_t) + \log (1 - D_{\phi_t}(\bm{h}_v)) 
    \right],
\end{equation}
where $D_{\phi_t}: \mathbb{R}^d \to [0,1]$ and $D_{\phi_v}$ are the modality discriminators, implemented as Multi-Layer Perceptrons (MLPs) with LeakyReLU activation functions. These discriminators are responsible for distinguishing between the visual and textual features, and they are trained to maximize the objective while the generator (disentanglement module) is trained to minimize it. Adversarial signals are delivered to the generator via a \textit{Gradient Reversal Layer (GRL)}, which compels $\bm{h}_t$ to be recognized as visual features by $D_{\phi_v}$, and $\bm{h}_v$ to be identified as textual features by $D_{\phi_t}$.

This adversarial process minimizes the Jensen-Shannon divergence between the distributions of the disentangled features and the opposing modality:
\begin{equation}
\mathcal{L}_{\text{adv}} = \text{JSD}(p(\bm{h}_t) \parallel p(\mathcal{V})) + \text{JSD}(p(\bm{h}_v) \parallel p(\mathcal{T})),
\end{equation}
where $p(\bm{h}_t)$ and $p(\bm{h}_v)$ are the distributions of the disentangled features, and $p(\mathcal{T})$ and $p(\mathcal{V})$ represent the distributions of the target modality. The adversarial equilibrium ensures that $\bm{h}_t$ and $\bm{h}_v$ reside in orthogonal modality subspaces, thereby achieving effective modality disentanglement.

\textbf{Mutual Information Minimization: Differentiable Disentanglement via CLUB.} To further enforce statistical independence between $\bm{h}_t$ and $\bm{h}_v$, a constraint on their mutual information (MI) is imposed using the \textit{Contrastive Log-ratio Upper Bound (CLUB)} estimator. The MI between two features $\bm{h}_t$ and $\bm{h}_v$ measures the amount of shared information between them, and minimizing it helps to ensure that the features are independent. Given paired disentangled features $(\bm{h}_t, \bm{h}_v)$, the MI is defined as:
\begin{equation}
    I(\bm{h}_t; \bm{h}_v) = \mathbb{E}_{p(\bm{h}_t, \bm{h}_v)} \left[ 
        \log \frac{p(\bm{h}_t|\bm{h}_v)}{p(\bm{h}_t)} 
    \right],
\end{equation}
where $p(\bm{h}_t|\bm{h}_v)$ is the conditional distribution of $\bm{h}_t$ given $\bm{h}_v$, and $p(\bm{h}_t)$ is the marginal distribution of $\bm{h}_t$. As the conditional distribution $p(\bm{h}_t|\bm{h}_v)$ is generally intractable, it is approximated by a variational distribution $q_\psi(\bm{h}_t|\bm{h}_v)$, which is parameterized as a Gaussian mixture model:
\begin{equation}
    I_{\text{CLUB}} = \mathbb{E}_{p(\bm{h}_t, \bm{h}_v)} [\log q_\psi(\bm{h}_t|\bm{h}_v)] - \mathbb{E}_{p(\bm{h}_t)p(\bm{h}_v)} [\log q_\psi(\bm{h}_t|\bm{h}_v)].
\end{equation}
Minimizing $I_{\text{CLUB}}$ encourages the disentangled features to become decorrelated in the probability density space, ensuring that $\bm{h}_t$ and $\bm{h}_v$ are independent. This approach allows for differentiable disentanglement, facilitating end-to-end training.

To optimize this objective, we adopt an alternating optimization strategy. Specifically, we update the variational distribution $q_\psi$ every $k$ steps to tighten the upper bound, then freeze $q_\psi$ to update the disentanglement module. This alternating optimization ensures stable convergence and effective decoupling of the features.

\textbf{Overall Training Framework.} The adversarial and mutual information minimization objectives jointly form the core of the training framework. By simultaneously optimizing the adversarial loss and mutual information minimization, we ensure that the disentangled features, $\bm{h}_t$ and $\bm{h}_v$, are both modality-agnostic and independent. The adversarial training prevents features from one modality from being mistaken for those of the other, while the CLUB-based mutual information minimization ensures that the features contain no mutual information, thereby guaranteeing effective disentanglement. In practice, the generator (disentanglement module) and discriminators are updated alternately. The disentanglement module is trained to minimize both the adversarial and mutual information losses, while the discriminators aim to maximize the adversarial loss. This min-max game framework, combined with the mutual information constraint, ensures the model learns to produce well-separated features that are effective for downstream tasks.

\subsection{Segmentation Model Text Modality Understanding and Training}
Sa2VA directly extracts latent representations from the multimodal large language model and integrates them into the segmentation model, effectively leveraging cross-modal predictions to supervise segmentation learning. Although real masks are used to train the segmentation model, the content provided to the model is not the true value. As a result, despite the involvement of the multimodal large language model output in training the segmentation model's decoder, the reliability of the information used for training remains uncertain. When the segmentation model receives prompts from the  large language model, these prompts are generated by the model itself and do not rely on real text information during training or inference.

To address this limitation, we introduce a pre-training phase that allows the segmentation model to learn text understanding without requiring additional real text inputs. During pre-training, we transform real text from the dataset into point-level information that can be directly processed by SAM2. The real text information is then combined with the pixel information processed by the segmentation model's encoder and fed into the model's decoder to generate the real text mask. The model is supervised by computing the pixel-wisecross-entropy loss \( \mathcal{L}_{\text{CE}} \) and Dice loss \( \mathcal{L}_{\text{DICE}} \) between the predicted text mask and the ground-truth mask, encouraging the model to effectively incorporate textual cues. Through this pre-training step, we develop a text decoder module within the segmentation model, enabling it to comprehend text.

In the first training phase, we unfreeze the text decoder to allow the model to learn and refine its ability to process text information. During the decoupling phase, the text decoder is frozen and used to complete the final training. The core of this function is to measure the orthogonality between the feature matrices $\mathbf{H}_{\text{label\_mask}}$ and $\mathbf{H}_{\text{gt\_mask}}$ using the squared \textbf{Frobenius} norm:
\begin{equation}	\mathcal{L}_{\text{ortho}} = \left\| \mathbf{H}_{\text{label\_mask}}^\top \mathbf{H}_{\text{gt\_mask}} \right\|_F^2, \quad \text{where} \quad 
\begin{cases}
\mathbf{H}_{\text{label\_mask}} = [\mathbf{h}_{\text{label\_mask}}^1, ..., \mathbf{h}_{\text{label\_mask}}^B]^\top, \\
\mathbf{H}_{\text{gt\_mask}} = [\mathbf{h}_{\text{gt\_mask}}^1, ..., \mathbf{h}_{\text{gt\_mask}}^B]^\top.
\end{cases}	\label{eq:ortho_loss}
\end{equation}

\subsection{Model Learning Scheme and Loss Functions}
The model uses visual modality information as point inputs, which are combined with the pixel-level features of the input image processed by the segmentation model's encoder. This combined data is then passed to the model's decoder to generate the visual prediction mask. The dataset includes the ground truth mask (\texttt{gt\_mask}) for each input image. The loss between the ground truth mask and the visual prediction mask (\texttt{visual}) is computed using pixel-level cross-entropy loss (\( \mathcal{L}_{\text{CE}} \)) and Dice loss (\( \mathcal{L}_{\text{DICE}} \)).
\begin{equation}
\mathcal{L}_{\text{intruction}} = \mathcal{L}_{\text{text}} + \mathcal{L}_{\text{masks}} , \quad\mathcal{L}_{\text{masks}} = \mathcal{L}_{\text{CE}} + \mathcal{L}_{\text{DICE}}.
\end{equation}

For text modality information, it undergoes the same processing as the visual modality via a linear layer. Both modalities share the same set of segmentation tokens (\texttt{seg\_token}), so after processing by the linear layer, the text modality information is formatted to match the visual modality. Both are now point-level data that SAM2 can handle, allowing identical processing for both modalities. However, to prevent the text modality’s linear layer from learning the same information as the visual modality, our model utilizes the dataset labels to pre-train SAM2’s text understanding module, enabling the model to comprehend text and assist in decoupling the text modality. 

The modality decoupler processes multimodal large language model (MLLM) outputs to extract textual features, which are subsequently decoded through a pre-trained text encoder. These decoded features undergo cross-modal fusion with the input image's pixel space to generate the final text segmentation mask prediction.

Our methodology explicitly avoids direct prediction-to-mask loss computation to prevent modality confusion. Rather than comparing text predictions with visual ground truth, we employ textual annotations from the dataset to generate reference text masks through joint processing of image pixels and text decoder outputs. These reference masks enable pixel-wise supervision using combined cross-entropy (\( \mathcal{L}_{\text{CE}} \)) and dice (\( \mathcal{L}_{\text{DICE}} \)) losses, enforcing strict text modality disentanglement while eliminating target ambiguity.
\begin{equation}
    \mathcal{L}_{\text{CE}} = \frac{1}{N} \sum_{i=1}^{N} \left( \hat{y}_i - y_i \right)^2,
\end{equation}
\begin{equation}
    \mathcal{L}_{\text{Dice}} = 1 - \frac{2 \sum_{i=1}^{N} y_i \hat{y}i}{\sum{i=1}^{N} y_i + \sum_{i=1}^{N} \hat{y}_i}.
\end{equation}
Our framework implements dual modality-specific supervision through distinct mask comparisons. The ground-truth visual mask supervises image feature extraction by computing reconstruction loss against predicted visual masks, while the textual ground-truth mask (\texttt{label\_mask}) enforces semantic alignment through pixel-wise cross-entropy and Dice losses against text predictions (\texttt{text}), strengthening the guidance provided by the text modality. This bifurcated approach enhances cross-modal coordination by maintaining separate yet complementary learning objectives for visual and textual processing streams.
Through this decoupling scheme, we obtain both the text prediction loss and the visual prediction loss. To fully leverage the decoupled text and visual information, we combine the text and visual prediction masks to generate the final prediction mask. The loss between the final prediction mask and the ground truth mask is computed to yield the final prediction loss. These three losses jointly supervise the learning of the model, enhancing its ability to leverage  information. By decoupling  information, we provide stronger guidance to the segmentation model, enabling it to fully utilize the  input for improved performance.

\subsection{Improved Dense Prompting for SAM2 via Self-Feedback Generation}

The \texttt{seg\_token} provides SAM2 with sparse semantic and visual prompts, limiting prompt refinement. To overcome this, we incorporate dense prompts by using the mask input as an additional prompt for SAM2. Inspired by the "self-feedback generation" technique~\cite{Madaan2023SelfRefineIR}, the model first generates a mask with the initial prompt, then reuses this mask as a dense prompt for refinement while keeping other inputs fixed. SAM2 predicts an improved mask based on this refined input, enabling iterative segmentation enhancement.
\begin{equation}
    \hat{M}_{t+1} = f(P_t, \hat{M}_t),
\end{equation}
where \( \hat{M}_t \) represents the predicted mask at iteration \( t \), and \( P_t \) represents the input prompt at iteration \( t \), which includes the previously generated mask as a dense prompt in subsequent iterations.

Our experiments show a single iteration achieves similar gains as multiple iterations while preserving inference speed. This approach improves mask quality for both video and image segmentation without adding parameters, effectively balancing accuracy and efficiency.

\section{Experiments}
\label{sec:experiment}

\subsection{Implementation Details}
Our framework enhances prior designs by integrating annotated data with specialized tags from a LLM. Specifically, we use InternVL2.5\_4b to generate textual question-answer pairs for interaction and latent signals for the segmentation model. SAM-2 is used as the segmentation model, receiving decoupled unimodal information from the output of the multi-modal large language model and pretrained with textual understanding from text annotations in the dataset. We train the model on four task datasets, including image and video question answering, as well as image and video segmentation, which together contain approximately 1.1 million image-text/video-text pairs. Training datasets include RefCOCO~\cite{yu2016modeling}, RefCOCO+\cite{yu2016modeling}, RefCOCOg\cite{yu2016modeling} for image segmentation, and MeVIS~\cite{ding2023mevis}, Ref-DAVIS17~\cite{seo2020urvos}, ReVOS~\cite{yan2024visa} for video segmentation. To retain question answering abilities, we use 66.5K LLaVA 1.5~\cite{liu2024llava} and 10K ChatUniVi~\cite{jin2024chat} data. We also incorporate Glamm\_data and Osprey-724k datasets to enhance fine-grained image-text alignment and large-scale foundational training. Training was conducted using the XTuner~\cite{contributors2023xtuner} codebase on eight NVIDIA H800 GPUs with a learning rate of 4e-5 and LoRA~\cite{hu2022lora} for LLM fine-tuning over 48 hours.

\subsection{Quantitative  Results}
\label{sec:result}

\textbf{Image/Video Segmentation Task.}  
Table~\ref{tab:REF} shows that our model with the proposed decoupling strategy achieves 82.6, 77.8, and 79.2 on RefCOCO~\cite{yu2016modeling}, RefCOCO+\cite{yu2016modeling}, and RefCOCOg\cite{yu2016modeling}, outperforming the baseline Sa2VA by 3.7, 6.1, and 5.1 points, respectively. This gain stems from explicitly transferring language model hidden signals to the segmentation model and employing a linear alignment layer to better integrate textual features, enhancing decoder training. On video segmentation benchmarks MeVIS~\cite{ding2023mevis}, Ref-DAVIS17~\cite{seo2020urvos}, and ReVOS~\cite{yan2024visa}, our model attains 46.7, 76.1, and 70.1, slightly surpassing Sa2VA. By decoupling visual and textual modalities, the segmentation model effectively handles heterogeneous inputs, boosting segmentation performance across tasks.

\setlength{\tabcolsep}{6.5pt}
\begin{table*}[ht]
	\centering
	\tiny 
	\caption{Comparison of model results across image/video segmentation tasks. Whether using the 1B or 4B model, our model outperforms baseline models in both image and video segmentation tasks.}
	\begin{tabular}{l|ccc|ccc}
		\toprule
		\textbf{Method} & \multicolumn{3}{c|}{Image Segmentation} & \multicolumn{3}{c}{Video Segmentation} \\
		 & RefCOCO ~\cite{yu2016modeling} & RefCOCO+ ~\cite{yu2016modeling} & RefCOCOg ~\cite{yu2016modeling} & MeViS ~\cite{ding2023mevis} & Ref-DAVIS17 ~\cite{seo2020urvos} & ReVOS ~\cite{yan2024visa} \\
		\midrule
        PixelLM-7B~\cite{ren2024pixellm} & 73.0 & 66.3 & 69.3 & - & - & - \\
        LaSagnA~\cite{wei2024lasagna} & 76.8 & 66.4 & 70.6 & - & - & - \\
		
		LISA-7B~\cite{lai2023lisa} & 74.1 & 62.4 & 66.4 & - & - & -\\
		GLaMM-7B~\cite{rasheed2024glamm} & 79.5 & 72.6 & 74.2 & - & - & - \\
		LLaVA-G-7B~\cite{liu2024llava} & 77.1 & 68.8 & 71.5 & - & - & - \\
		
        GSVA-13B~\cite{xia2024gsva} & 79.2 & 70.3 & 75.7 & - & - & - \\
        OMG-LLaVA-7B~\cite{zhang2024omg} & 78.0 & 69.1 & 72.9 & - & - & - \\
        
        VISA-13B~\cite{yan2024visa} & 72.4 & 59.8 & 65.5 & 44.5 & 70.4 & 50.9 \\
		\midrule
        Sa2VA-1B ~\cite{yuan2025sa2va} & 77.4 & 69.9 & 72.3 & 41.7 & 72.3 & 47.6 \\
		Sa2VA-4B ~\cite{yuan2025sa2va} & 78.9 & 71.7 & 74.1 & 46.2 & 73.8 & 53.2 \\
		\rowcolor{gray!20} 
        \textbf{DeSa2VA-1B (Ours)} & \textbf{78.9} & \textbf{75.7} & \textbf{72.5} & \textbf{45.1} & \textbf{74.4} & \textbf{48.3} \\
        \rowcolor{gray!20}
		\textbf{DeSa2VA-4B (Ours)} & \textbf{82.6} & \textbf{77.8} & \textbf{79.2} & \textbf{47.2} & \textbf{77.0} & \textbf{54.4} \\
		\bottomrule
	\end{tabular}
	\label{tab:REF}
\end{table*}

\textbf{Image/Video Question Answering Task.}  
As shown in Table~\ref{tab:GCG}, the 4-B model trained with the decoupling strategy on LLaVA-1.5 (665K)~\cite{liu2024llava} achieves a score of 73.3 on the SEED-Bench~\cite{li2023seed} image question answering benchmark, and also scores 50.4 on the Video-MME~\cite{fu2024video} video question answering benchmark, comparable to the baseline Sa2VA. These results are consistent with the design objective of the decoupling algorithm, which aims to enhance the segmentation model by decoupling and transferring the hidden outputs of the  language model. Since the  language model handles the question answering tasks, the decoupling strategy does not interfere with its understanding or reasoning capabilities. These results demonstrate that our method preserves the language model’s QA performance while improving the segmentation-specific components.

\begin{table*}[ht]
	\centering
	\scriptsize 
	\caption{Comparison of results of different methods across QA tasks and GCG tasks.}
	\begin{tabular}{l|c|c|c}
		\toprule
		\textbf{Method} & Image Chat & Video Chat & GCG \\
		 & SEED-Bench~\cite{li2023seed} & Video-MME~\cite{fu2024video} & GCG ~\cite{rasheed2024glamm} \\
		\midrule
		LLAVA-1.5-13B~\cite{liu2024llava} & 70.1 & - & - \\
		Video-LLaVA-7B~\cite{lin2023video} & - & 39.9 & - \\
		LLaMA-VID-7B~\cite{li2024llama} & 59.9 & - & - \\
        mPLUG-Owl3-8B~\cite{ye2024mplug} & - & 53.5 & - \\
        InternVL2-8B~\cite{chen2024far} & 76.2 & 54.0 & - \\
		GLaMM-7B~\cite{rasheed2024glamm} & - & - & 28.9 \\
        OMG-LLaVA-7B~\cite{zhang2024omg} & 56.5 & - & 29.9 \\
		\midrule
        Sa2VA-1B ~\cite{yuan2025sa2va} & 64.8 & 39.9 & 23.8 \\
        Sa2VA-4B ~\cite{yuan2025sa2va} & 73.3 & 50.4 & 28.2 \\
       
        \rowcolor{gray!20} 
        \textbf{DeSa2VA-1B (Ours)} & \textbf{65.1} & \textbf{39.9} & \textbf{24.1}  \\ 
		\rowcolor{gray!20} 
        \textbf{DeSa2VA-4B (Ours)} &  \textbf{73.3}  & \textbf{50.4}&  \textbf{28.7}  \\
		\bottomrule
	\end{tabular}
	\label{tab:GCG}
\end{table*}

\textbf{Results on GCG Validation Set.}  
In the Grounded Caption Generation (GCG) task~\cite{rasheed2024glamm}, we evaluate the model’s ability to align images and text. Region localization accuracy is measured using the mean Intersection over Union (mIoU) of segmentation masks, and text accuracy is validated with BLEU and CIDEr scores. Cross-modal recall is also introduced to assess alignment between phrases and masks. As shown in Table~\ref{tab:GCG}, after information decoupling with enhanced prompts, the 1B and 4B models achieve scores of 24.1 and 28.7, outperforming the baselines. These results demonstrate strong segmentation and text generation performance, with stable cross-modal associations in complex scenes. This experiment confirms the effectiveness of multi-modal decoupling and enhanced prompts for fine-grained image-text alignment.

\subsection{Qualitative Results}

\begin{figure}[ht]
	\centering
\includegraphics[width=1\textwidth]{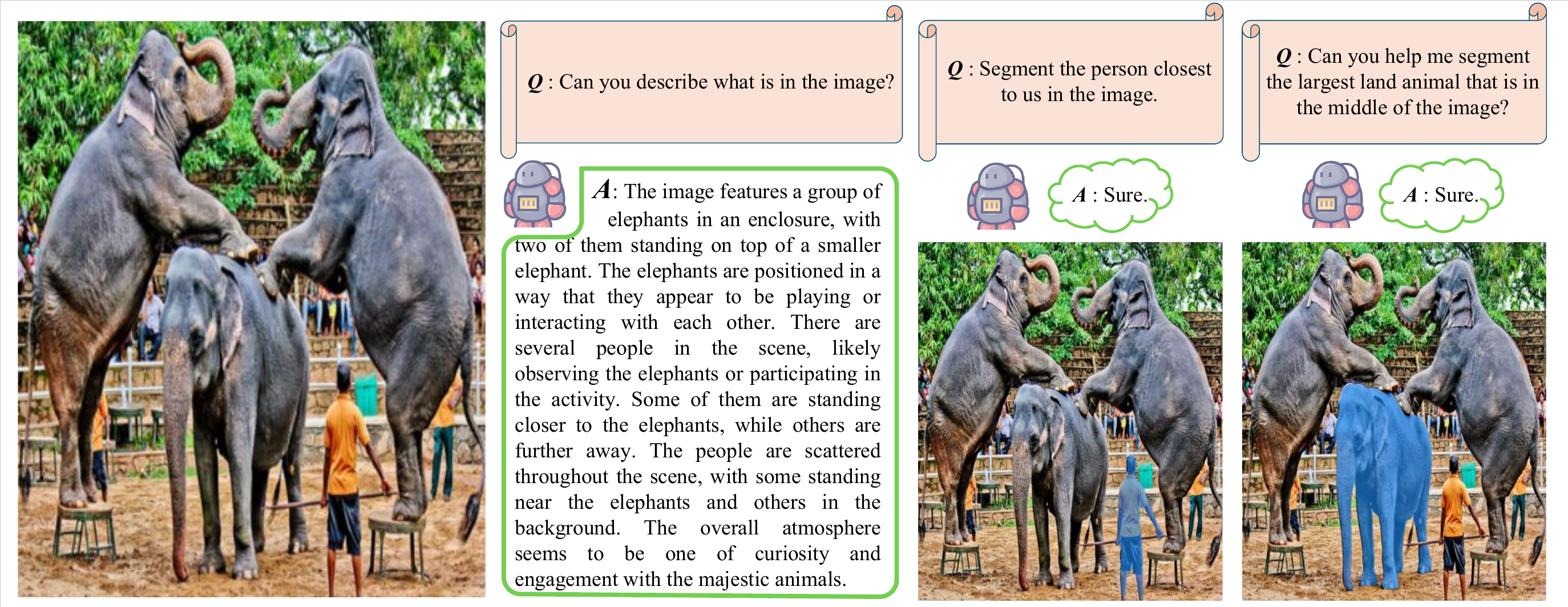}
	\caption{ \textbf{Question-Answering and Segmentation Results.} The left side presents the input image along with the model's accurate description, while the right side displays the segmentation results, demonstrating the model's capability to fulfill segmentation requirements.} 
	\label{fig:example}
\end{figure}

Figure~\ref{fig:example} illustrates our model's joint performance on visual question answering and semantic segmentation tasks. The experiment utilizes a web-sourced image processed through our unified framework, which simultaneously performs visual reasoning to generate both textual responses and pixel-level segmentation masks. Notably, the model demonstrates dynamic adaptation to textual cues in user queries for segmentation tasks.

\subsection{Ablation Studies}
\textbf{Dataset Ablation Experiment.}
The datasets include image QA, image segmentation, video QA, and video segmentation, each targeting specific capabilities. In the ablation study (Table~\ref{tab:wo}), removing image segmentation data causes a 74\% drop in segmentation accuracy, while excluding video segmentation data only reduces performance by 5–10\%, indicating image segmentation alone is sufficient. Textual training enhances decoding, enabling strong generalization with limited image segmentation data. Text-based QA performance remains stable across dataset variations, reflecting effective modality decoupling. Notably, models without video segmentation data outperform those trained on both image and video data in static segmentation, suggesting potential interference from video data that merits further investigation.
\setlength{\tabcolsep}{8pt}
\begin{table*}[ht]
	\centering
	\tiny 
	\caption{The model was trained on four tasks, image/video segmentation and image/video understanding. In the ablation experiment, we trained the model on three tasks at a time and compared results.}
	\begin{tabular}{l|ccc|cc|c}
		\toprule
		\textbf{Data} & \multicolumn{3}{c|}{Image Segmentation} & \multicolumn{2}{c|}{Video Segmentation} & GCG \\
		& RefCOCO ~\cite{yu2016modeling} & RefCOCO+ ~\cite{yu2016modeling} & RefCOCOg ~\cite{yu2016modeling} & MeViS ~\cite{ding2023mevis} & Ref-DAVIS17 ~\cite{seo2020urvos} & GCG ~\cite{rasheed2024glamm} \\
		\midrule
		All data & 82.6 & 77.8 & 79.2 & 47.2 & 77.0 & 28.7 \\
		w/o Image QA & 80.0 & 75.2 & 77.0 & 46.7 & 77.2 & 27.9 \\
		w/o Image Segmentation & 21.5 & 20.4 & 25.6 & 35.8 & 40.6 & 13.6 \\
        w/o Video QA & 83.0 & 77.9 & 79.7 & 44.2 & 75.4 & 28.8 \\
		w/o Video Segmentation & 77.3 & 76.1 & 76.8 & 40.5 & 72.7 & 26.5 \\
		\bottomrule
	\end{tabular}
	\label{tab:wo}
\end{table*}

\textbf{Ablation of Training Process.}
Table~\ref{tab:cmp} shows that our decoupling-enhanced prompt method achieves the best segmentation results when SAM-2 is pre-trained on real textual labels. Without this pre-training, the decoupled model still outperforms baselines on RefCOCO, RefCOCO+, RefCOCOg, and Ref-DAVIS17, demonstrating strong generalization. In contrast, a baseline trained only for text understanding underperforms in visual tasks, revealing a trade-off between modalities. These findings underscore the importance of modality-specific pre-training to fully leverage enhanced cues. Without text understanding pre-training, segmentation suffers from poor textual interpretation; applying our decoupling prompt further boosts performance, confirming its necessity for multimodal segmentation.

\setlength{\tabcolsep}{7.2pt}
\begin{table*}[ht]
	\centering
	\tiny 
	\caption{Comparison of model results with and without pre-trained SAM-2. "pre-train" represents text understanding pre-training, while "w/o pre-train" means without text understanding pre-training. }
	\begin{tabular}{l|ccc|cc|c}
		\toprule
		\textbf{Method} & \multicolumn{3}{c|}{Image Segmentation} & \multicolumn{2}{c|}{Video Segmentation} & GCG \\
		& RefCOCO ~\cite{yu2016modeling} & RefCOCO+ ~\cite{yu2016modeling} & RefCOCOg ~\cite{yu2016modeling} & MeViS ~\cite{ding2023mevis} & Ref-DAVIS17 ~\cite{seo2020urvos} & GCG ~\cite{rasheed2024glamm} \\
		\midrule
		Sa2VA-4B (w/o pre-train) & 78.9 & 71.7 & 74.1 & 46.2 & 73.8 & 28.2 \\
        Sa2VA-4B (pre-train) & 73.4 & 67.9 & 69.6 & 40.1 & 70.4 & 25.2 \\
		DeSa2VA-4B (w/o pre-train) & 79.2 & 75.3 & 76.6 & 45.9 & 74.8 & 28.5 \\
		\rowcolor{gray!20}
		\textbf{DeSa2VA-4B (Ours)} & \textbf{82.6} & \textbf{77.8} & \textbf{79.2} &\textbf{47.2} & \textbf{77.0} & \textbf{28.7} \\
		\bottomrule
	\end{tabular}
	\label{tab:cmp}
\end{table*}

\textbf{Reception of Prompt Types in Segmentation Models.}
As shown in Table ~\ref{tab:masks}, for both the proposed and baseline models, the segmentation model with mask-level input prompts outperforms the one using only point-level input prompts. This demonstrates the feasibility and general applicability of enhancing the model with mask-based reprompting.

\begin{table*}[ht]
	\centering
	\small 
	\caption{Ablation study on cue types for the segmentation model when trained solely on the segmentation task dataset. “Mask reprompt” refers to using the output mask as a reprompt, whereas “w/o Mask reprompt” indicates freezing the mask input module.}
	\begin{tabular}{l|ccc}
		\toprule
		\textbf{Method} & RefCOCO ~\cite{yu2016modeling} & RefCOCO+ ~\cite{yu2016modeling} & RefCOCOg ~\cite{yu2016modeling} \\
		\midrule
		Sa2VA-1B (w/o Masks\_reprompt)~\cite{yuan2025sa2va} & 76.08 & 72.32 & 75.73 \\
        Sa2VA-1B (Masks\_reprompt)~\cite{yuan2025sa2va} & 78.02 & 72.41 & 76.02 \\
		DeSa2VA-1B (w/o Masks\_reprompt) & 77.56 & 72.61 & 75.94 \\
		\rowcolor{gray!20}
		\textbf{DeSa2VA-1B (Masks\_reprompt)} & \textbf{78.89} & \textbf{72.72} & \textbf{76.53} \\
		\bottomrule
	\end{tabular}
	\label{tab:masks}
\end{table*}

\section{Conclusion}
\label{sec:conclusion}
In this work, we propose DeSa2VA, a novel framework that integrates large language and segmentation models via an information decoupling module to produce unified multi-modal outputs. By separating textual and visual cues, and leveraging dedicated text understanding pre-training, our method enables precise interpretation and effective fusion, leading to superior performance on image/video segmentation and question answering tasks. Ablation studies validate the strong generalization and effectiveness of our decoupling strategy.

We also identify that video segmentation training can degrade image segmentation performance. Future work will explore temporal decoupling to better handle sequential information.
In conclusion, our work effectively decouples textual and visual modalities through a novel prompting and pre-training strategy, significantly enhancing segmentation performance across diverse image and video tasks while demonstrating strong generalization and practical applicability.

\section{Appendix}
\label{sec:ov}
This part provides additional content that complements the main text, including following aspects:
\begin{itemize}
    \item \textbf{Appendix A} gives more experiment results on our DeSa2VA models and more tasks.
    
    \item \textbf{Appendix B} gives visual examples of Sa2VA on various tasks.
    
\end{itemize}

\subsection{Appendix A: More Experiment Results}
\label{sec:A}
\textbf{More Complete Ablation Experiments on Reception of Prompt Types in Segmentation Models.} 
In Table~\ref{tab:masks} of the main paper, we train our 1B model solely on the segmentation datasets RefCOCO \cite{yu2016modeling}, RefCOCO+ \cite{yu2016modeling}, and RefCOCOg \cite{yu2016modeling}, and conduct ablation experiments only on these three image segmentation tasks. Tables~\ref{tab:REFV} and~\ref{tab:GCG2} present the ablation results of our 1B and 4B models when jointly trained on multiple datasets across different tasks.We train on four types of datasets including the image segmentation datasets RefCOCO \cite{yu2016modeling}, RefCOCO+ \cite{yu2016modeling}, and RefCOCOg \cite{yu2016modeling}, the image question answering dataset 66.5K LLaVA-1.5 \cite{liu2024llava}, the video segmentation datasets MeVIS~\cite{ding2023mevis}, Ref-DAVIS17~\cite{seo2020urvos}, and ReVOS~\cite{yan2024visa}, and the video question answering dataset 10K ChatUniVi \cite{jin2024chat}, along with additional Glamm\_data and Osprey-724k datasets.

\setlength{\tabcolsep}{6.5pt}
\begin{table*}[ht]
	\centering
	\tiny 
	\caption{More complete ablation experiments on reception of prompt types in segmentation models. This table presents the results of the segmentation tasks.}
	\begin{tabular}{l|ccc|cc}
		\toprule
		\textbf{Method} & \multicolumn{3}{c}{Image Segmentation} & \multicolumn{2}{c}{Video Segmentation} \\
		 & RefCOCO ~\cite{yu2016modeling} & RefCOCO+ ~\cite{yu2016modeling} & RefCOCOg ~\cite{yu2016modeling} & MeViS ~\cite{ding2023mevis} & Ref-DAVIS17 ~\cite{seo2020urvos} \\
		\midrule
        Sa2VA-1B (w/o Masks\_reprompt)~\cite{yuan2025sa2va} & 77.4 & 69.9 & 72.3 & 41.7 & 72.3 \\
        Sa2VA-1B (Masks\_reprompt)~\cite{yuan2025sa2va} & 78.5 & 70.1 & 72.3 & 43.0 & 73.1 \\
		DeSa2VA-1B (w/o Masks\_reprompt) & 78.5 & 74.6 & 72.3 & 44.2 & 74.0 \\
		DeSa2VA-1B (Masks\_reprompt) & 78.9 & 75.7 & 72.5 & 45.1 & 74.4 \\
        \midrule
        Sa2VA-4B (w/o Masks\_reprompt)~\cite{yuan2025sa2va} & 78.9 & 71.7 & 74.1 & 46.2 & 73.8 \\
        Sa2VA-4B (Masks\_reprompt)~\cite{yuan2025sa2va} & 80.9 & 71.8 & 75.0 & 47.0 & 74.0 \\
		DeSa2VA-4B (w/o Masks\_reprompt) & 81.3 & 77.1 & 78.0 & 47.0 & 76.7 \\
        DeSa2VA-4B (Masks\_reprompt) & 82.6 & 77.8 & 79.2 & 47.2 & 77.0 \\
		\bottomrule
	\end{tabular}
	\label{tab:REFV}
\end{table*}

As shown in Table~\ref{tab:REFV}, when the model enables the mask input module and applies output mask reprompting, both the 1B and 4B model variants outperform their respective baseline models and the DeSa2VA models without output mask reprompt on the image and video segmentation tasks. Similarly, baseline models with mask input enabled and output mask reprompt also surpass the original baseline models. In contrast, as shown in Table~\ref{tab:GCG2}, on image and video question answering tasks, model performance remains unaffected by the output mask reprompt. 

These results demonstrate that mask reprompting has broad applicability in segmentation tasks. Moreover, due to the model’s decoupling of visual and textual information, reprompting at the segmentation module does not negatively impact the model’s question answering capabilities.

\begin{table*}[ht]
	\centering
	\scriptsize 
	\caption{More complete ablation experiments on reception of prompt types in segmentation models. This table presents the results of the question answering and joint tasks.}
	\begin{tabular}{l|c|c|c}
		\toprule
		\textbf{Method} & Image Chat & Video Chat & GCG \\
		 & SEED-Bench~\cite{li2023seed} & Video-MME~\cite{fu2024video} & GCG ~\cite{rasheed2024glamm} \\
		\midrule
        Sa2VA-1B (w/o Masks\_reprompt)~\cite{yuan2025sa2va} & 64.8 & 39.9 & 23.8 \\
        Sa2VA-1B (Masks\_reprompt)~\cite{yuan2025sa2va} & 65.0 & 40.1 & 23.9 \\
		DeSa2VA-1B (w/o Masks\_reprompt) & 65.0 & 39.9 & 23.9 \\
		DeSa2VA-1B (Masks\_reprompt) & 65.1 & 39.9 & 24.1 \\
        \midrule
        Sa2VA-4B (w/o Masks\_reprompt)~\cite{yuan2025sa2va} & 73.3 & 50.4 & 28.2 \\
        Sa2VA-4B (Masks\_reprompt)~\cite{yuan2025sa2va} & 73.3 & 50.4 & 28.5 \\
		DeSa2VA-4B (w/o Masks\_reprompt) & 73.2 & 50.4 & 28.5 \\
        DeSa2VA-4B (Masks\_reprompt) & 73.3 & 50.4 & 28.7 \\
		\bottomrule
	\end{tabular}
	\label{tab:GCG2}
\end{table*}

\textbf{More Dataset Ablation Experiment.}
Table~\ref{tab:wo2} presents ablation experiments on image and video question answering tasks using the SEED-Bench~\cite{li2023seed} and Video-MME~\cite{fu2024video} datasets, with joint training on different multiple task datasets. When any single dataset is omitted from the combined training set, which includes image question answering, image segmentation, video question answering, and video segmentation datasets, no performance drop is observed on either image or video question answering tests. This demonstrates the strong generalization capability of our model.

\setlength{\tabcolsep}{8pt}
\begin{table*}[ht]
	\centering
	\tiny 
	\caption{Dataset ablation experiment on image and video QA tasks.}
	\begin{tabular}{l|c|c}
		\toprule
		\textbf{Method} & Image Chat & Video Chat \\
		& SEED-Bench~\cite{li2023seed} & Video-MME~\cite{fu2024video} \\
		\midrule
		All data & 73.3 & 50.4 \\
		w/o Image QA & 70.5 & 50.1 \\
		w/o Image Segmentation & 73.4 & 50.5 \\
        w/o Video QA & 72.9 & 49.8 \\
		w/o Video Segmentation & 73.3 & 51.0 \\
		\bottomrule
	\end{tabular}
	\label{tab:wo2}
\end{table*}

\textbf{More Ablation Experiment of Training Process.}
In the main text, we present ablation experiments on segmentation model text understanding pretraining using the 4B models. Table~\ref{tab:cmp1} shows the ablation results of segmentation model text understanding pretraining on the 1B models. After decoupling visual and textual information, our pretrained model shows significant improvement compared to its non-pretrained counterpart. In contrast, for the baseline model, which does not decouple visual and textual information, pretraining on textual understanding does not necessarily have a positive impact on performance. These results indicate that, regardless of model size, text understanding pretraining for the segmentation model is an indispensable component in DeSa2VA.

\setlength{\tabcolsep}{7.2pt}
\begin{table*}[ht]
	\centering
	\tiny 
	\caption{Ablation experiment of training process on 1B models.}
	\begin{tabular}{l|ccc|cc|c}
		\toprule
		\textbf{Method} & \multicolumn{3}{c|}{Image Segmentation} & \multicolumn{2}{c|}{Video Segmentation} & GCG \\
		& RefCOCO ~\cite{yu2016modeling} & RefCOCO+ ~\cite{yu2016modeling} & RefCOCOg ~\cite{yu2016modeling} & MeViS ~\cite{ding2023mevis} & Ref-DAVIS17 ~\cite{seo2020urvos} & GCG ~\cite{rasheed2024glamm} \\
		\midrule
		Sa2VA-1B (w/o pre-train) & 77.4 & 69.9 & 72.3 & 41.7 & 72.3 & 23.8\\
        Sa2VA-1B (pre-train) & 72.8 & 70.1 & 70.5 & 39.2 & 69.7 & 22.9 \\
		DeSa2VA-1B (w/o pre-train) & 78.0 & 74.9 & 72.5 & 40.9 & 73.1 & 24.0 \\
		DeSa2VA-1B (Ours) & 78.9 & 75.7 & 72.5 & 45.1 & 74.4 & 24.1 \\
		\bottomrule
	\end{tabular}
	\label{tab:cmp1}
\end{table*}

\subsection{Appendix B: Visualization Results}
\label{sec:B}

\textbf{Visualization of Video Segmentation Results under Complex Semantic Tasks.}  
\begin{figure}
    \centering
    \includegraphics[width= 1 \linewidth]{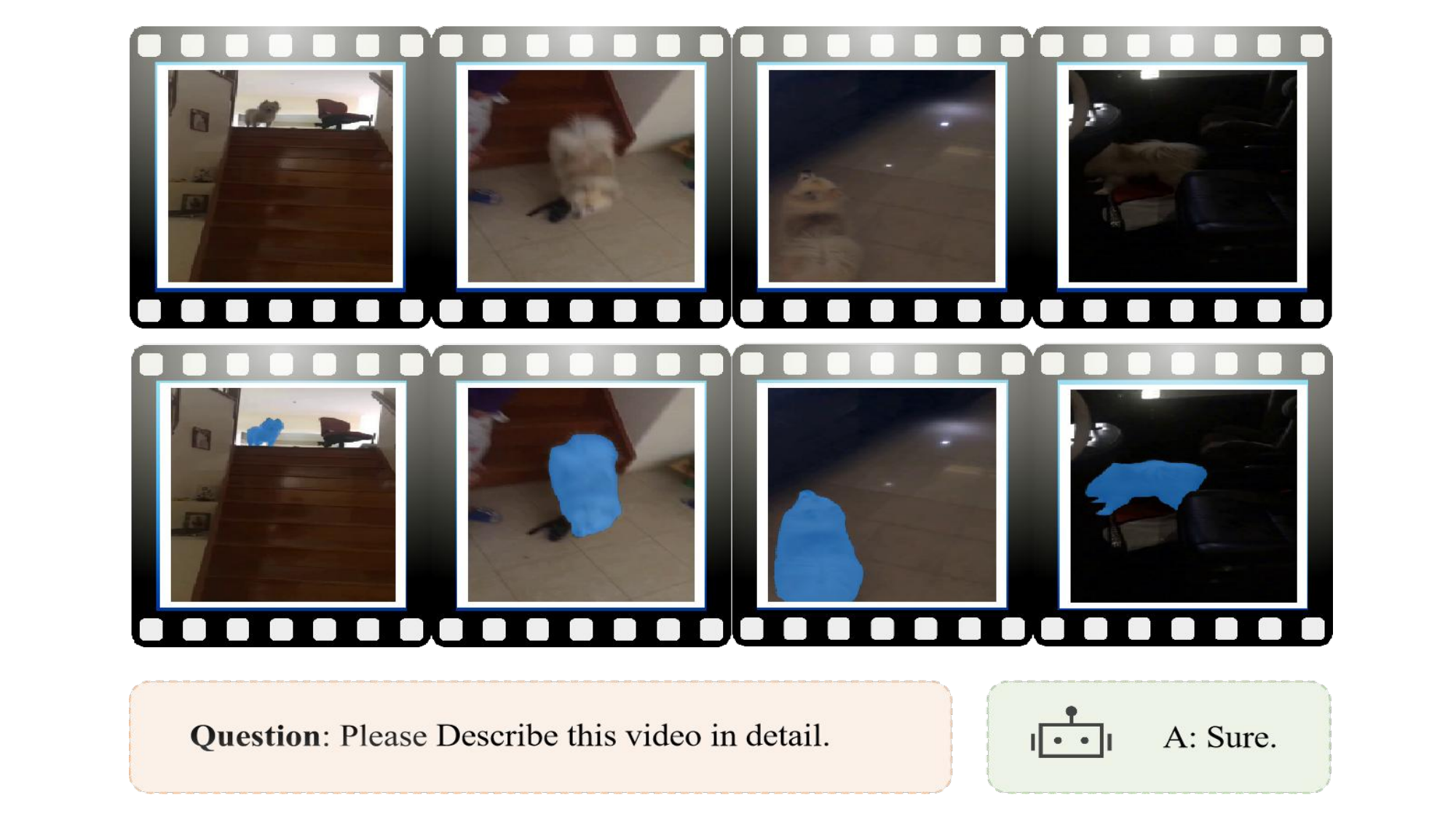}
    \caption{Visualization of the model’s segmentation performance under complex semantic and low-light scenarios.}
    \label{fig:enter-label1}
\end{figure}
Figure ~\ref{fig:enter-label1} illustrates the video segmentation results of the DeSa2VA model under complex semantic conditions. The task is to "Segment the fast-moving object in the video." As shown in Figure ~\ref{fig:enter-label1}, the model accurately segments the fastest moving dog in the video. Additionally, this example includes challenges such as occluded objects and dim lighting, including a partially visible dog and a dog in shadow. Our model performs robustly in handling these situations.

\textbf{Visualization of Video Understanding Results under Complex Scenes.}  
\begin{figure}
    \centering
    \includegraphics[width=1\linewidth]{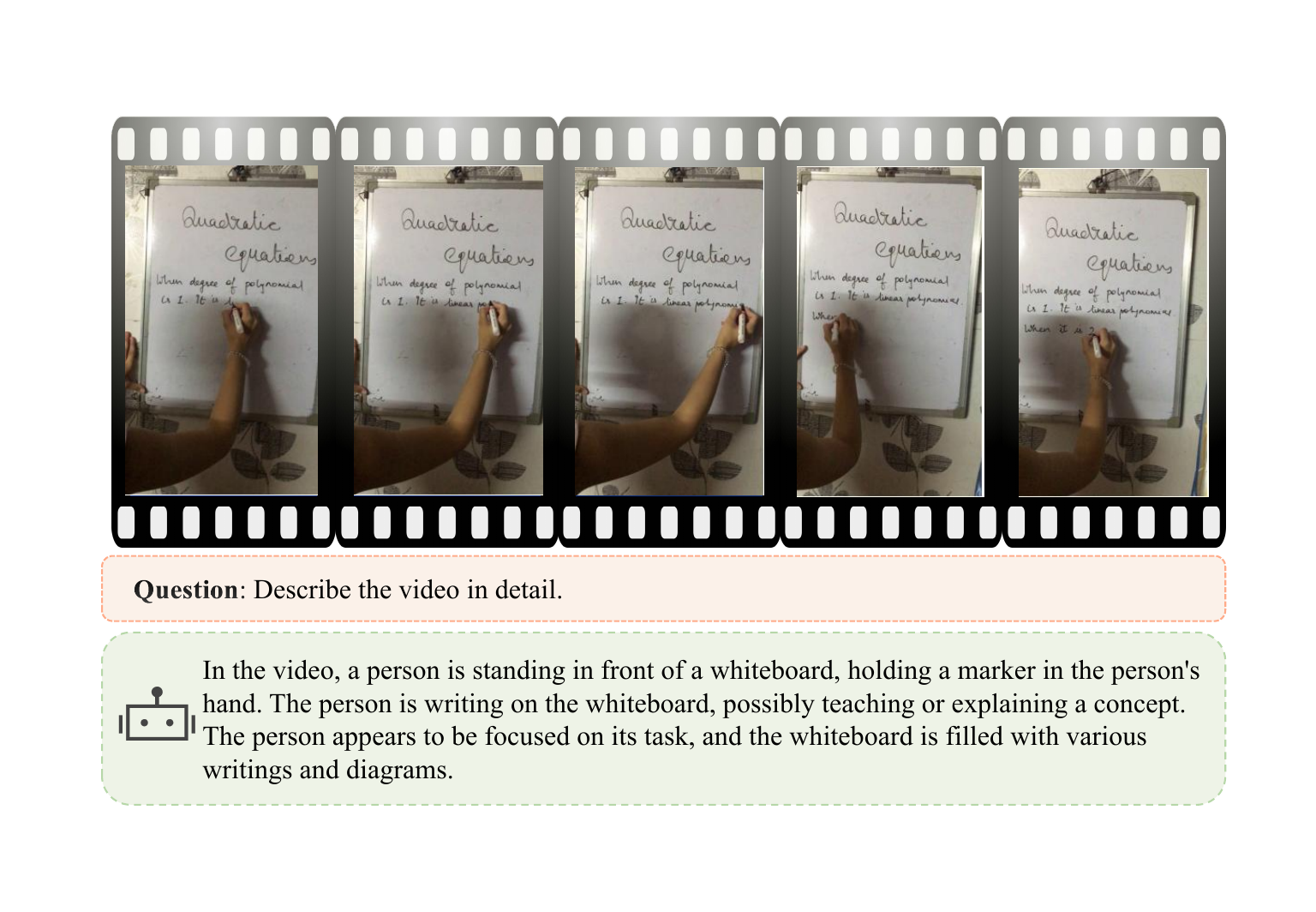}
    \caption{Visualization of the model’s segmentation performance under complex and partial scene tasks.}
    \label{fig:enter-label2}
\end{figure}
Figure ~\ref{fig:enter-label2} presents the video understanding results of the DeSa2VA model in complex scenes. The input video contains a hand writing complex text on a whiteboard. Our model successfully recognizes that the video depicts a person writing on a whiteboard, likely teaching or explaining a concept. This demonstrates the model’s strong capability in video understanding tasks involving complex scenarios.

\textbf{Visualization of Joint Segmentation and Understanding Results in Complex Scenes.}  
\begin{figure}
    \centering
    \includegraphics[width=1\linewidth]{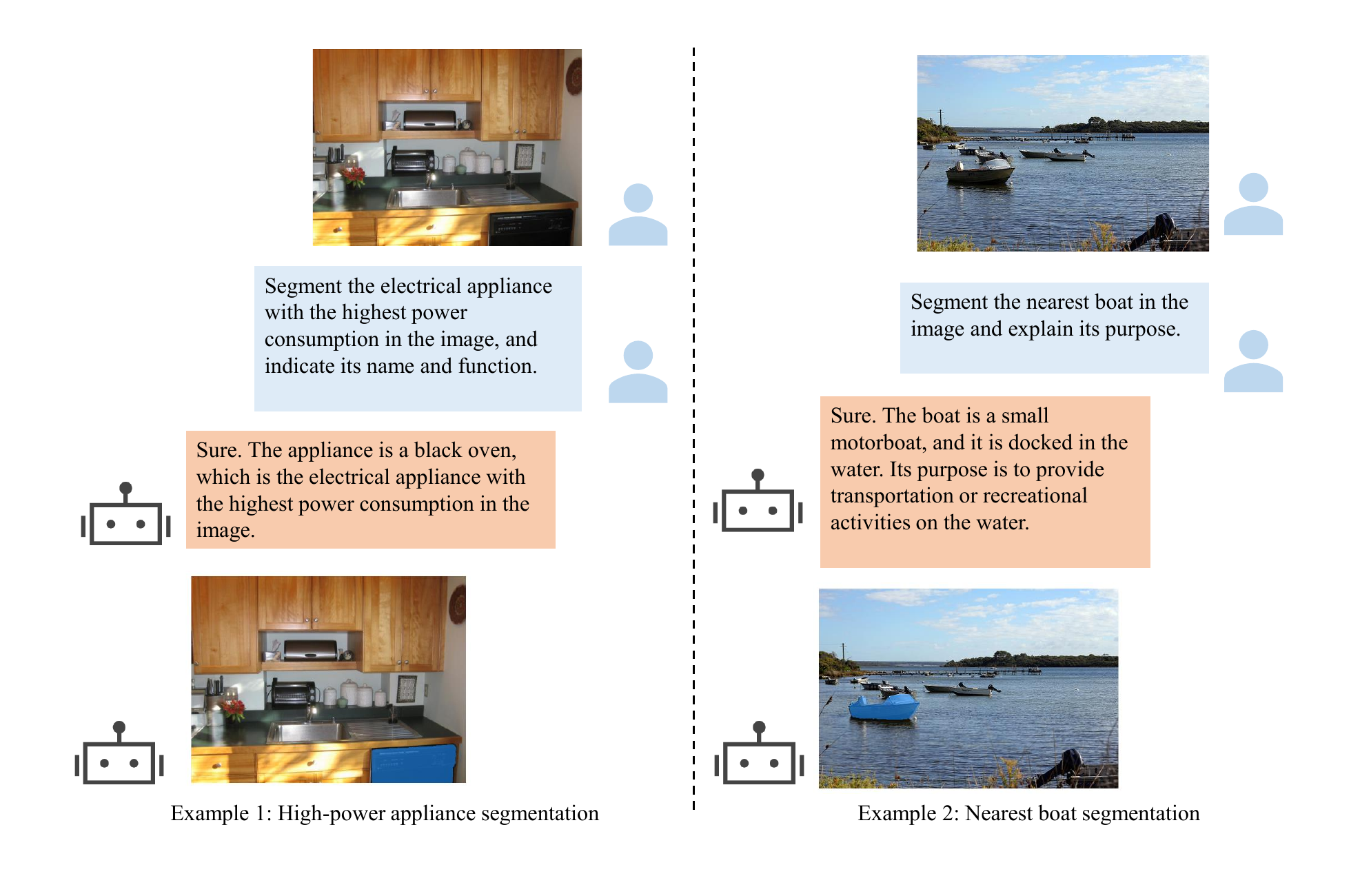}
    \caption{Visualization of the model’s results on the joint segmentation and question answering tasks.}
    \label{fig:enter-label3}
\end{figure}
Figure ~\ref{fig:enter-label3} shows the results of the DeSa2VA model on a task combining segmentation and understanding within a single instruction. The task is shown in the picture, and the results are shown in Figure ~\ref{fig:enter-label3}. The model accurately completes both the segmentation and the understanding (Q\&A) tasks. This indicates the model’s excellent performance in joint tasks while reducing computational cost and improving efficiency.

\bibliographystyle{plain}
\bibliography{Ref} 

\begin{thebibliography}{10}

\bibitem{bai2023qwen}
Jinze Bai, Shuai Bai, Shusheng Yang, Shijie Wang, Sinan Tan, Peng Wang, Junyang Lin, Chang Zhou, and Jingren Zhou.
\newblock Qwen-vl: A versatile vision-language model for understanding, localization, text reading, and beyond.
\newblock {\em arXiv preprint arXiv:2308.12966}, 2023.

\bibitem{brown2020language}
Tom Brown, Benjamin Mann, Nick Ryder, Melanie Subbiah, Jared~D Kaplan, Prafulla Dhariwal, Arvind Neelakantan, Pranav Shyam, Girish Sastry, Amanda Askell, and et~al.
\newblock Language models are few-shot learners.
\newblock In {\em NeurIPS}, 2020.

\bibitem{chen2023llava}
Wei-Ge Chen, Irina Spiridonova, Jianwei Yang, Jianfeng Gao, and Chunyuan Li.
\newblock Llava-interactive: An all-in-one demo for image chat, segmentation, generation and editing.
\newblock {\em arXiv preprint arXiv:2311.00571}, 2023.

\bibitem{chen2024far}
Zhe Chen, Weiyun Wang, Hao Tian, Shenglong Ye, Zhangwei Gao, Erfei Cui, Wenwen Tong, Kongzhi Hu, Jiapeng Luo, Zheng Ma, et~al.
\newblock How far are we to gpt-4v? closing the gap to commercial multimodal models with open-source suites.
\newblock {\em Science China Information Sciences}, 67(12):220101, 2024.

\bibitem{chen2024internvl}
Zhe Chen, Jiannan Wu, Wenhai Wang, Weijie Su, Guo Chen, Sen Xing, Muyan Zhong, Qinglong Zhang, Xizhou Zhu, Lewei Lu, et~al.
\newblock Internvl: Scaling up vision foundation models and aligning for generic visual-linguistic tasks.
\newblock In {\em Proceedings of the IEEE/CVF Conference on Computer Vision and Pattern Recognition}, pages 24185--24198, 2024.

\bibitem{contributors2023xtuner}
XTuner Contributors.
\newblock Xtuner: A toolkit for efficiently fine-tuning llm, 2023.

\bibitem{dang2025mupa}
Jisheng Dang, Huilin Song, Junbin Xiao, Bimei Wang, Han Peng, Haoxuan Li, Xun Yang, Meng Wang, and Tat-Seng Chua.
\newblock Mupa: Towards multi-path agentic reasoning for grounded video question answering.
\newblock {\em arXiv preprint arXiv:2506.18071}, 2025.

\bibitem{dang2025reinforcing}
Jisheng Dang, Jingze Wu, Teng Wang, Xuanhui Lin, Nannan Zhu, Hongbo Chen, Wei-Shi Zheng, Meng Wang, and Tat-Seng Chua.
\newblock Reinforcing video reasoning with focused thinking.
\newblock {\em arXiv preprint arXiv:2505.24718}, 2025.

\bibitem{dang2025synpo}
Jisheng Dang, Yizhou Zhang, Hao Ye, Teng Wang, Siming Chen, Huicheng Zheng, Yulan Guo, Jianhuang Lai, and Bin Hu.
\newblock Synpo: Synergizing descriptiveness and preference optimization for video detailed captioning.
\newblock {\em arXiv preprint arXiv:2506.00835}, 2025.

\bibitem{ding2023mevis}
Henghui Ding, Chang Liu, Shut~ing He, Xu~dong Jiang, and Chen~Change Loy.
\newblock Mevis: A large-scale benchmark for video segmentation with motion expressions.
\newblock In {\em ICCV}, 2023.

\bibitem{dong2024mastering}
Xiaoyi Dong, Pan Zhang, Yuhang Zang, Yuhang Cao, Bin Wang, Linke Ouyang, Songyang Zhang, Haodong Duan, Maosong Cao, Wenwei Zhang, Yining Li, Hang Yan, Yang Gao, Xinyue Zhang, Wei Li, Jingwen Li, Wenhai Wang, Kai Chen, Conghui He, Xingcheng Zhang, Jiefeng Dai, Yu~Qiao, Dahua Lin, and Jiaqi Wang.
\newblock Mastering free-form text-image composition and comprehension in vision-language large model.
\newblock {\em arXiv preprint arXiv:2401.16420}, 2024.

\bibitem{fu2023mme}
Chaoyou Fu, Peixian Chen, Yunhang Shen, Yulei Qin, Mengdan Zhang, Xu~Lin, Jinrui Yang, Xiawu Zheng, Ke~Li, Xing Sun, Yunsheng Wu, and Rongrong Ji.
\newblock Mme: A comprehensive evaluation benchmark for multimodal large language models.
\newblock {\em arXiv preprint arXiv:2306.13394}, 2023.

\bibitem{fu2024video}
Chaoyou Fu, Yuhan Dai, Yongdong Luo, Lei Li, Shuhuai Ren, Renrui Zhang, Zihan Wang, Chenyu Zhou, Yunhang Shen, Mengdan Zhang, et~al.
\newblock Video-mme: The first-ever comprehensive evaluation benchmark of multi-modal llms in video analysis.
\newblock {\em arXiv preprint arXiv:2405.21075}, 2024.

\bibitem{guo2023openvis}
Pinxue Guo, Tony Huang, Peiyang He, Xuefeng Liu, Tianjun Xiao, Zhaoyu Chen, and Wenqiang Zhang.
\newblock Openvis: Open-vocabulary video instance segmentation.
\newblock {\em arXiv preprint arXiv:2305.16835}, 2023.

\bibitem{hu2022lora}
Edward~J Hu, Yelong Shen, Phillip Wallis, Zeyuan Allen-Zhu, Yuanzhi Li, Shen Wang, Lu~Wang, and Weizhu Chen.
\newblock Lora: Low-rank adaptation of large language models.
\newblock In {\em ICLR}, 2022.

\bibitem{huang2024vtimelm}
Bin Huang, Xin Wang, Hong Chen, Zihan Song, and Wenu Zhu.
\newblock Vtimelm: Empower llm to grasp video moments.
\newblock In {\em CVPR}, 2024.

\bibitem{huang2025reason3d}
Kuan-Hui Huang, Xiangtai Li, Lu~Qi, Shuicheng Yan, and Ming-Hsuan Yang.
\newblock Reason3d: Searching and reasoning 3d segmentation via large language model.
\newblock In {\em 3DV}, 2025.

\bibitem{huang2020pixel}
Zhicheng Huang, Zhaoyang Zeng, Bei Liu, Dongmei Fu, and Jianlong Fu.
\newblock Pixel-bert: Aligning image pixels with text by deep multi-modal transformers.
\newblock {\em arXiv preprint arXiv:2004.00849}, 2020.

\bibitem{hudson2019gqa}
Drew~A Hudson and Christopher~D Manning.
\newblock Gqa: A new dataset for real-world visual reasoning and compositional question answering.
\newblock In {\em CVPR}, 2019.

\bibitem{hwang2021video}
Sukjun Hwang, Miran Heo, Seung~Wug Oh, and Seon~Joo Kim.
\newblock Video instance segmentation using inter-frame communication transformers.
\newblock In {\em NeurIPS}, 2021.

\bibitem{jin2024chat}
Peng Jin, Ryuichi Takanobu, Wancai Zhang, Xiaochun Cao, and Li~Yuan.
\newblock Chat-univi: Unified visual representation empowers large language models with image and video understanding.
\newblock In {\em Proceedings of the IEEE/CVF Conference on Computer Vision and Pattern Recognition}, pages 13700--13710, 2024.

\bibitem{khoreva2018video}
Anna Khoreva, Anna Rohrbach, and Bernt Schiele.
\newblock Video object segmentation with language referring expressions.
\newblock In {\em ACCV}, 2018.

\bibitem{kirillov2023segment}
Alexander Kirillov, Eric Mintun, Nikhila Ravi, Hanzi Mao, Chloe Rolland, Laura Gustafson, Tete Xiao, Spencer Whitehead, Alexander~C Berg, Wan-Yen Lo, et~al.
\newblock Segment anything.
\newblock In {\em Proceedings of the IEEE/CVF International Conference on Computer Vision}, pages 4015--4026, 2023.

\bibitem{lai2024lisa}
Xin Lai, Zhiotao Tian, Yukang Chen, Yanwei Li, Yuhui Yuan, Shu Liu, and Jiaya Jia.
\newblock Lisa: Reasoning segmentation via large language model.
\newblock In {\em CVPR}, 2024.

\bibitem{lai2023lisa}
Xin Lai, Zhuotao Tian, Yukang Chen, Yanwei Li, Yuhui Yuan, Shu Liu, and Jiaya Jia.
\newblock Lisa: Reasoning segmentation via large language model.
\newblock {\em arXiv preprint arXiv:2308.00692}, 2023.

\bibitem{li2024llava-onevision}
Bo~Li, Yuanhan Zhang, Dong Guo, Renrui Zhang, Feng Li, Hao Zhang, Kaichen Zhang, Yanwei Li, Ziwei Liu, and Chuyuan Li.
\newblock Llava-onevision: Easy visual task transfer.
\newblock {\em arXiv preprint arXiv:2408.03326}, 2024.

\bibitem{li2023seed}
Bohao Li, Rui Wang, Guangzhi Wang, Yuying Ge, Yixiao Ge, and Ying Shan.
\newblock Seed-bench: Benchmarking multimodal llms with generative comprehension.
\newblock {\em arXiv preprint arXiv:2307.16125}, 2023.

\bibitem{li2023blip}
Junnan Li, Dongxu Li, Silvio Savarese, and Steven Hoi.
\newblock Blip-2: Bootstrapping language-image pre-training with frozen image encoders and large language models.
\newblock In {\em International conference on machine learning}, pages 19730--19742. PMLR, 2023.

\bibitem{li2022blip}
Junnan Li, Dongxu Li, Caiming Xiong, and Steven Hoi.
\newblock Blip: Bootstrapping language-image pre-training for unified vision-language understanding and generation.
\newblock In {\em ICML}, 2022.

\bibitem{li2023tube-link}
Xiangtai Li, Haobo Yuan, Wenwei Zhang, Guangliang Cheng, Jingmiu Pang, and Chen~Change Loy.
\newblock Tube-link: A flexible cross tube baseline for universal video segmentation.
\newblock In {\em ICCV}, 2023.

\bibitem{li2024llama}
Yanwei Li, Chengyao Wang, and Jiaya Jia.
\newblock Llama-vid: An image is worth 2 tokens in large language models.
\newblock In {\em European Conference on Computer Vision}, pages 323--340. Springer, 2024.

\bibitem{lin2023video}
Bin Lin, Bin Zhu, Yang Ye, Munan Ning, Peng Jin, and Li~Yuan.
\newblock Video-llava: Learning united visual representation by alignment before projection.
\newblock {\em arXiv preprint arXiv:2311.10122}, 2023.

\bibitem{liu2024visual}
Haotian Liu, Chunyuan Li, Qingyang Wu, and Yong~Jae Lee.
\newblock Visual instruction tuning.
\newblock {\em Advances in neural information processing systems}, 36, 2024.

\bibitem{liu2024llava}
Shilong Liu, Hao Cheng, Haotian Liu, Hao Zhang, Feng Li, Tianhe Ren, Xueyan Zou, Jianwei Yang, Hang Su, Jun Zhu, et~al.
\newblock Llava-plus: Learning to use tools for creating multimodal agents.
\newblock In {\em European Conference on Computer Vision}, pages 126--142. Springer, 2024.

\bibitem{luqi2024generalizable}
LuQi, Yi-Wen Chen, Lehan Yang, Tiancheng Shen, Xiangtai Li, Weidong Guo, Yu~Xu, and Ming-Hsuan Yang.
\newblock Generalizable entity grounding via assistance of large language model.
\newblock {\em arXiv preprint arXiv:2402.02555}, 2024.

\bibitem{Madaan2023SelfRefineIR}
Aman Madaan, Niket Tandon, Prakhar Gupta, Skyler Hallinan, Luyu Gao, Sarah Wiegreffe, Uri Alon, Nouha Dziri, Shrimai Prabhumoye, Yiming Yang, Sean Welleck, Bodhisattwa~Prasad Majumder, Shashank Gupta, Amir Yazdanbakhsh, and Peter Clark.
\newblock Self-refine: Iterative refinement with self-feedback.
\newblock {\em ArXiv}, abs/2303.17651, 2023.

\bibitem{rasheed2024glamm}
Hanooma Rasheed, Muhammad Maaz, Sahal Shaji, Abdelrahman Shaker, Salman Khan, Hisham Cholakkal, Rao~M. Anwer, Eric Xing, Ming-Hsuan Yang, and Fahad~S. Al-Ansari.
\newblock Glamm: Pixel grounding large multimodal model.
\newblock In {\em CVPR}, 2024.

\bibitem{ravi2024sam}
Nikhila Ravi, Valentin Gabourie, Yuan-Ting Hu, Ronghang Hu, Chaitanya Ryali, Tengyu Ma, Haitham Khedr, Roman Rädle, Chloe Rolland, Laura Gustafson, et~al.
\newblock Sam 2: Segment anything in images and videos.
\newblock {\em arXiv preprint arXiv:2408.00714}, 2024.

\bibitem{ren2024pixellm}
Zhongwei Ren, Zhicheng Huang, Yunchao Wei, Yao Zhao, Dongmei Fu, Jiashi Feng, and Xiaojie Jin.
\newblock Pixellm: Pixel reasoning with large multimodal model.
\newblock In {\em CVPR}, 2024.

\bibitem{seo2020urvos}
Seoung Seo, Joon-Young Lee, and Bohyung Han.
\newblock Urvos: Unified referring video object segmentation network with a large-scale benchmark.
\newblock In {\em ECCV}, 2020.

\bibitem{touvron2023llama2}
Hugo Touvron, Louis Martin, Kevin Stone, Peter Albert, Amjad Almahairi, Yasmine Babaei, Nikolay Bashlykov, Soumya Batra, Prajjwal Bhargava, Shruti Bhosale, et~al.
\newblock Llama 2: Open foundation and fine-tuned chat models.
\newblock {\em arXiv preprint arXiv:2307.09288}, 2023.

\bibitem{wei2024lasagna}
Cong Wei, Haoxian Tan, Yujie Zhong, Yujiu Yang, and Lin Ma.
\newblock Lasagna: Language-based segmentation assistant for complex queries.
\newblock {\em arXiv preprint arXiv:2404.08506}, 2024.

\bibitem{xia2024gsva}
Zhuofan Xia, Dongchen Han, Yizeng Han, Xuran Pan, Shiji Song, and Gao Huang.
\newblock Gsva: Generalized segmentation via multimodal large language models.
\newblock In {\em Proceedings of the IEEE/CVF Conference on Computer Vision and Pattern Recognition}, pages 3858--3869, 2024.

\bibitem{yan2023universal}
Bin Yan, Yi~Jiang, Jiannan Wu, Dong Wang, Zehuan Yuan, Ping Luo, and Huchuan Lu.
\newblock Universal instance perception as object discovery and retrieval.
\newblock In {\em CVPR}, 2023.

\bibitem{yan2024visa}
Cilin Yan, Haochen Wang, Shilin Yan, Xiaolong Jiang, Yao Hu, Guoliang Kang, Weidi Xie, and Efstratios Gavves.
\newblock Visa: Reasoning video object segmentation via large language models.
\newblock {\em arXiv preprint arXiv:2407.11325}, 2024.

\bibitem{ye2024mplug}
Jiabo Ye, Haiyang Xu, Haowei Liu, Anwen Hu, Ming Yan, Qi~Qian, Ji~Zhang, Fei Huang, and Jingren Zhou.
\newblock mplug-owl3: Towards long image-sequence understanding in multi-modal large language models.
\newblock {\em arXiv preprint arXiv:2408.04840}, 2024.

\bibitem{yu2016modeling}
Lichen Yu, Patrick Poisson, Shan Yang, Alexander~C Berg, and Tamara~L Berg.
\newblock Modeling context in referring expressions.
\newblock In {\em ECCV}, 2016.

\bibitem{yuan2025sa2va}
H.~Yuan, X.~Li, T.~Zhang, Z.~Huang, S.~Xu, S.~Ji, Y.~Tong, L.~Qi, J.~Feng, and M.~Yang.
\newblock Sa2va: Marrying sam2 with llava for dense grounded understanding of images and videos.
\newblock In {\em Proceedings of the IEEE/CVF Conference on Computer Vision and Pattern Recognition (CVPR)}, 2025.

\bibitem{zhang2024omg}
Tao Zhang, Xiangtai Li, Hao Fei, Haobo Yuan, Shengqiong Wu, Shunping Ji, Change~Loy Chen, and Shuicheng Yan.
\newblock Omg-llava: Bridging image-level, object-level, pixel-level reasoning and understanding.
\newblock In {\em NeurIPS}, 2024.

\bibitem{zhou2023rethinking}
Hao Zhou, Tiancheng Shen, Xu~Yang, Hai Huang, Xiangtai Li, Lu~Qi, and Ming-Hsuan Yang.
\newblock Rethinking evaluation metrics of open-vocabulary segmentation.
\newblock {\em arXiv preprint arXiv:2311.03352}, 2023.

\bibitem{zhu2022instance}
Feng Zhu, Zongxin Yang, Xin Yu, Yi~Yang, and Yunchao Wei.
\newblock Instance as identity: A generic online paradigm for video instance segmentation.
\newblock In {\em ECCV}, 2022.

\end{thebibliography}

\clearpage

\clearpage

\end{document}